\documentclass[11pt,a4paper]{article}
\usepackage[hyperref]{emnlp-ijcnlp-2019}
\usepackage{times}
\usepackage{latexsym}

\usepackage{multirow}
\usepackage{float}
\usepackage{graphicx}
\usepackage{amsmath}
\usepackage{amsfonts}
\usepackage{amssymb}
\usepackage{todonotes}

\usepackage{graphicx}
\usepackage{subcaption}

\usepackage[utf8]{inputenc}
\usepackage[T1]{fontenc}
\usepackage[russian,english]{babel}
\usepackage{booktabs, tabularx}

\usepackage{url}

\aclfinalcopy 

\title{Context-Aware Monolingual Repair for Neural Machine Translation}

  \author{Elena Voita$^{1,2}$ \quad Rico Sennrich$^{4,3}$ \quad Ivan Titov$^{3,2}$\bigskip\\
  $^1$Yandex, Russia \quad 
  $^2$University of Amsterdam, Netherlands \\
  $^3$University of Edinburgh, Scotland  \quad
  $^4$University of Zurich, Switzerland \\
  {\tt lena-voita@yandex-team.ru} \\ {\tt sennrich@cl.uzh.ch} \quad {\tt ititov@inf.ed.ac.uk}
}

\date{}

\begin{document}
\maketitle

\begin{abstract}

Modern sentence-level NMT systems often produce plausible translations of isolated sentences.  However, when put in context, these translations may end up being inconsistent with each other. 
We propose a monolingual DocRepair model to correct inconsistencies between sentence-level translations. DocRepair performs automatic post-editing on a sequence of sentence-level translations, refining translations of sentences in context of each other. For training, the DocRepair model requires only monolingual document-level data in the target language. It is trained as a monolingual sequence-to-sequence model that maps inconsistent groups of sentences into consistent ones. The consistent groups come from the original training data; the inconsistent groups are obtained by sampling round-trip translations for each isolated sentence. We show that this approach successfully imitates inconsistencies we aim to fix: using contrastive evaluation, we show large improvements in the translation of several contextual phenomena in an English$\to$Russian translation task, as well as improvements in the BLEU score. We also conduct a human evaluation and show a strong preference of the annotators to corrected translations over the baseline ones. Moreover, we analyze which discourse phenomena are hard to capture using monolingual data only.\footnote{The code and data sets (including round-trip translations) are available at \url{https://github.com/lena-voita/good-translation-wrong-in-context}.}

\end{abstract}

\section{Introduction}

Machine translation has made remarkable progress, and studies claiming it to reach a human parity are starting to appear~\cite{hassan2018achieving}. However, when evaluating translations of the whole documents rather than isolated sentences, human raters show a stronger preference for human over machine translation~\cite{laubli-D18-1512}. These findings emphasize the need to shift towards context-aware machine translation both from modeling and evaluation perspective.

Most previous work on context-aware NMT assumed that either all the bilingual data is available at the document level~\cite{jean_does_2017,wang_exploiting_2017,tiedemann_neural_2017,bawden2017,voita18,maruf-haffari:2018:Long,agrawal2018,kuang-etal-2018-modeling,miculicich-EtAl:2018:EMNLP} or at least its fraction~\cite{voita-etal-2019-good}. But in practical scenarios, document-level parallel data is often scarce, which is one of the challenges when building a context-aware system.

We introduce an approach to context-aware machine translation using only monolingual document-level data. In our setting, a separate monolingual sequence-to-sequence model (DocRepair) is used to correct sentence-level translations of adjacent sentences.
The key idea is to use monolingual data to imitate typical inconsistencies between context-agnostic translations of isolated sentences. The DocRepair model is trained to map inconsistent groups of sentences into consistent ones. The consistent groups come from the original training data; the inconsistent groups are obtained by sampling round-trip translations for each isolated sentence.

To validate the performance of our model, we use three kinds of evaluation: the BLEU score, contrastive evaluation of translation of several discourse phenomena \cite{voita-etal-2019-good}, and human evaluation. We show strong improvements for all metrics.

We analyze which discourse phenomena are hard to capture using monolingual data only. Using contrastive test sets for targeted evaluation of several contextual phenomena, we compare the performance of the models trained on round-trip translations and genuine document-level parallel data.  
Among the four phenomena in the test sets we use (deixis, lexical cohesion, VP ellipsis and ellipsis which affects NP inflection) we find VP ellipsis to be the hardest phenomenon to be captured using round-trip translations. 

Our key contributions are as follows:
\begin{itemize}
    \item we introduce the first approach to context-aware machine translation using only monolingual document-level data;
    \item our approach shows substantial improvements in translation quality as measured by BLEU, targeted contrastive evaluation of several discourse phenomena and human evaluation;
    \item we show which discourse phenomena are hard to capture using monolingual data only.
\end{itemize}

\section{Our Approach: Document-level Repair}
\label{sect:our_approach}

We propose a monolingual DocRepair model to correct inconsistencies between sentence-level translations of a context-agnostic MT system. It does not use any states of a trained MT model whose outputs it corrects and therefore can in principle be trained to correct translations from any black-box MT system. 

\begin{figure}[t!]
\center{\includegraphics[scale=0.65]{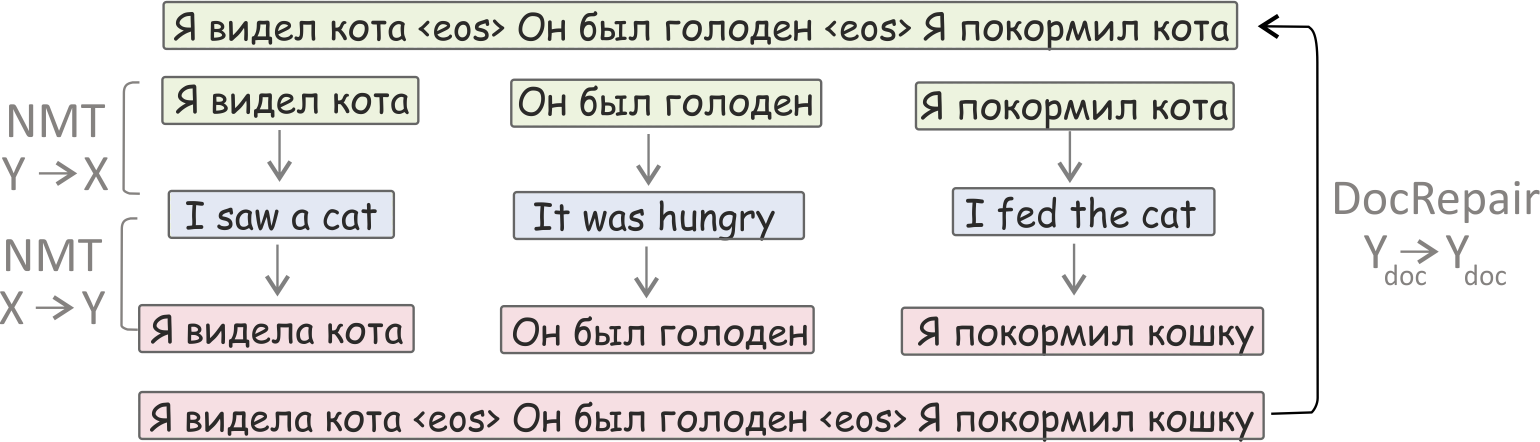}}
\caption{Training procedure of DocRepair. First, round-trip translations of individual sentences are produced to form an inconsistent text fragment (in the example, both genders of the speaker and the cat became inconsistent). Then, a repair model is trained to produce an original text from the inconsistent one.}
\label{fig:train_doc_repair}
\end{figure}

The DocRepair model requires only monolingual document-level data in the target language. It is a monolingual sequence-to-sequence model that maps inconsistent groups of sentences into consistent ones. Consistent groups come from monolingual document-level data. To obtain inconsistent groups, each sentence in a group is replaced with its round-trip translation produced in isolation from context. More formally, forming a training minibatch for the DocRepair model involves the following steps (see also Figure~\ref{fig:train_doc_repair}):
\begin{enumerate}
    \item sample several groups of sentences from the monolingual data;
    \item for each sentence in a group, (i) translate it using a target-to-source MT model, (ii) sample a translation of this back-translated sentence in the source language using a source-to-target MT model;
    \item using these round-trip translations of isolated sentences, form an inconsistent version of the initial groups;
    \item use inconsistent groups as input for the DocRepair model, consistent ones as output.
\end{enumerate}

\begin{figure}[t!]
\center{\includegraphics[scale=0.65]{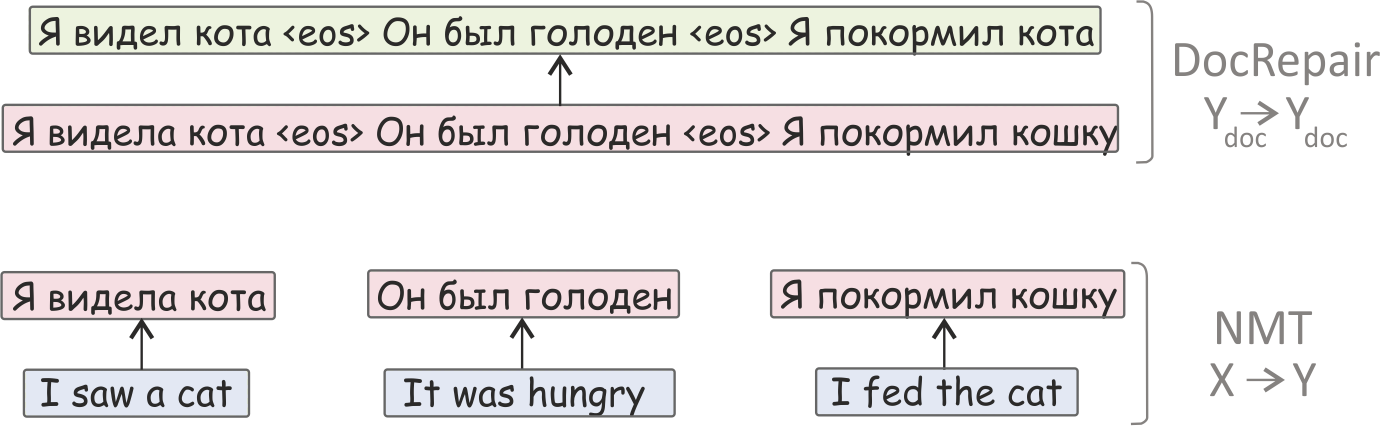}}
\caption{The process of producing document-level translations at test time is two-step: (1) sentences are translated independently using a sentence-level model, (2) DocRepair model corrects translation of the resulting text fragment.}
\label{fig:test_doc_repair}
\end{figure}

At test time, the process of getting document-level translations is two-step (Figure~\ref{fig:test_doc_repair}):
\begin{enumerate}
    \item produce translations of isolated sentences using a context-agnostic MT model;
    \item apply the DocRepair model to a sequence of context-agnostic translations to correct inconsistencies between translations.
\end{enumerate}

In the scope of the current work, the DocRepair model is the standard sequence-to-sequence Transformer. Sentences in a group are concatenated using a reserved token-separator between sentences.\footnote{In preliminary experiments, we observed that this performs better than concatenating sentences without a separator.} The Transformer is trained to correct these long inconsistent pseudo-sentences into consistent ones. The token-separator is then removed from corrected translations.

\section{Evaluation of Contextual Phenomena}
\label{sect:evaluation_context_awareness}

We use contrastive test sets for evaluation of discourse phenomena for English-Russian by~\citet{voita-etal-2019-good}. These test sets allow for testing different kinds of phenomena which, as we show, can be captured from monolingual data with varying success. In this section, we provide test sets statistics and briefly describe the tested phenomena. For more details, the reader is referred to~\citet{voita-etal-2019-good}. 

\subsection{Test sets}

There are four test sets in the suite.
Each test set contains contrastive examples. It is specifically designed to test the ability of a system to adapt to contextual information and handle the phenomenon under consideration. Each test instance consists of a true example (a sequence of sentences and their reference translation from the data) and several contrastive translations which differ from the true one only in one specific aspect. All contrastive translations are correct and plausible translations at the sentence level, and only context reveals the inconsistencies between them. The system is asked to score each candidate translation, and we compute the system accuracy as the proportion of times the true translation is preferred to the contrastive ones. Test set statistics are shown in Table~\ref{tab:testset_stat}. The suites for deixis and lexical cohesion are split into development and test sets, with 500 examples from each used for validation purposes and the rest for testing. Convergence of both consistency scores on these development sets and BLEU score on a general development set are used as early stopping criteria in models training. For ellipsis, there is no dedicated development set, so we evaluate on all the ellipsis data and do not use it for development.

\begin{table}[t!]
\centering
\begin{tabular}{lcccc}
\toprule
& & \multicolumn{3}{c}{distance}\\
  & total & 1 & 2 & 3 \\ 
 \cmidrule{2-5}
\bf deixis   & 3000 & 1000 & 1000 & 1000\\
\bf lex. cohesion   & 2000 & \phantom{0}855 & \phantom{0}630 & \phantom{0}515\\
\bf ellipsis (infl.) & \phantom{0}500\\
\bf ellipsis (VP) & \phantom{0}500\\
\bottomrule
\end{tabular}\textbf{}
\caption{Size of test sets: total number of test instances and with regard to the distance between sentences requiring consistency (in the number of sentences).
For ellipsis, the two sets correspond to whether a model has to predict correct NP inflection, or correct verb sense (VP ellipsis). 
} \label{tab:testset_stat}
\end{table}

\subsection{Phenomena overview}

\textbf{Deixis} Deictic words or phrases, are referential expressions whose denotation depends on context. This includes personal deixis (``I'', ``you''), place deixis (``here'', ``there''), and discourse deixis, where parts of the discourse are referenced (``that's a good question''). 
The test set examples are all related to person deixis, specifically the T-V distinction between informal and formal you (Latin ``tu'' and ``vos'') in the Russian translations, and test for consistency in this respect.

\textbf{Ellipsis} Ellipsis is the omission from a clause of one or more words that are nevertheless understood in the context of the remaining elements. In machine translation, elliptical constructions in the source language pose a problem in two situations. First, if the target language does not allow the same types of ellipsis, requiring the elided material to be predicted from context. Second, if the elided material affects the syntax of the sentence. For example, in Russian the grammatical function of a noun phrase, and thus its inflection,  may depend on the elided verb, or, conversely, the verb inflection may depend on the elided subject.

There are two different test sets for ellipsis. One contains 
examples where a morphological form of a noun group in the last sentence can not be understood without context beyond the sentence level (``ellipsis (infl.)'' in Table~\ref{tab:testset_stat}). Another includes cases of verb phrase ellipsis in English, which does not exist in Russian, thus requires predicting the verb when translating into Russian (``ellipsis (VP)'' in Table~\ref{tab:testset_stat}).

\textbf{Lexical cohesion} The test set focuses on reiteration of named entities. Where several translations of a named entity are possible, a model has to prefer consistent translations over inconsistent ones.

\section{Experimental Setup}
\label{sect:exp_setup}

\subsection{Data preprocessing}
\label{sect:data_preprocessing}

We use the publicly available OpenSubtitles2018 corpus~\cite{LISON18.294} for English and Russian.
For a fair comparison with previous work, we train the baseline MT system on the data released by~\citet{voita-etal-2019-good}. Namely, our MT system is trained on 6m instances. These are sentence pairs with a relative time overlap of subtitle frames between source  and target language subtitles of at least $0.9$. 

We gathered 30m groups of 4 consecutive sentences as our monolingual data. We used only documents not containing groups of sentences from general development and test sets as well as from contrastive test sets. The main results we report are for the model trained on all 30m fragments. 

We use the tokenization provided by the corpus and use {\tt multi-bleu.perl}\footnote{\url{https://github.com/moses-smt/mosesdecoder/tree/master/scripts/generic}} on lowercased data to compute BLEU score. We use beam search with a beam of 4.

Sentences were encoded using byte-pair encoding~\cite{sennrich-bpe}, with source and target vocabularies of about 32000 tokens.
Translation pairs were batched together by approximate sequence length. Each training batch contained a set of translation pairs containing approximately 15000\footnote{This can be reached by using several of GPUs or by accumulating the gradients for several batches and then making an update.} source tokens. It has been shown that Transformer's performance depends heavily on batch size~\cite{training-tips-transformer}, and we chose a large batch size to ensure the best performance. In training context-aware models, for early stopping we use both convergence in BLEU score on the general development set and scores on the consistency development sets. After training, we average the 5 latest checkpoints.

\subsection{Models}
\label{sect:model_parameters}

The baseline model, the model used for back-translation, and the DocRepair model are all {\it Transformer base} models~\cite{attention-is-all-you-need}.  More precisely, the number of layers is $N=6$ with $h = 8$ parallel attention layers, or heads. The dimensionality of input and output is $d_{model} = 512$, and the inner-layer of a feed-forward networks has dimensionality $d_{ff}=2048$. We use regularization as described in~\citet{attention-is-all-you-need}.

As a second baseline, we use the two-pass CADec model~\cite{voita-etal-2019-good}. The first pass produces  sentence-level translations. The second pass takes both the first-pass translation and representations of the context sentences as input and returns contextualized translations. CADec requires document-level parallel training data, while DocRepair only needs monolingual training data.

\subsection{Generating round-trip translations}
\label{sect:generate_round_trip}

On the selected 6m instances we train sentence-level translation models in both directions.
To create training data for DocRepair, we proceed as follows.  The Russian monolingual data is first translated into English, using the Russian$\to$English model and beam search with beam size  of 4. Then, we use the English$\to$Russian model to sample translations with temperature of $0{.}5$. For each sentence, we precompute 20 sampled translations and randomly choose one of them when forming a training minibatch for DocRepair. Also, in training, we replace each token in the input with a random one with the probability of $10\%$.

\subsection{Optimizer}
\label{sect:optimizer}

As in~\citet{attention-is-all-you-need},
we use the Adam optimizer~\cite{adam-optimizer}, the parameters are $\beta_1 = 0{.}9$, $\beta_2 = 0{.}98$ and $\varepsilon = 10^{-9}$. We vary the learning rate over the course of training using the formula:
\begin{multline*}
l_{rate}=scale\cdot \min(step\_num^{-0.5},\\ step\_num\cdot warmup\_steps^{-1.5}),
\end{multline*}
where $warmup\_steps = 16000$ and $scale=4$.

\section{Results}

\subsection{General results}

The BLEU scores are provided in Table~\ref{tab:bleu_scores} (we evaluate translations of 4-sentence fragments). To see which part of the improvement is due to fixing agreement between sentences rather than simply sentence-level post-editing, we train the same repair model at the sentence level. Each sentence in a group is now corrected separately, then they are put back together in a group. One can see that most of the improvement  comes from accounting for extra-sentential dependencies. DocRepair outperforms the baseline and CADec by 0{.}7 BLEU, and its sentence-level repair version by 0{.}5 BLEU.

\begin{table}[t!]
\centering
\begin{tabular}{lc}
\toprule
 \bf model & \bf BLEU \\ 
 \cmidrule{1-2}
baseline   &  33{.}91 \\
CADec & 33{.}86 \\
sentence-level repair & 34{.}12 \\
DocRepair   &  \bf 34{.}60 \\
\bottomrule
\end{tabular}\textbf{}
\caption{BLEU scores. For CADec, the original implementation was used.}
\label{tab:bleu_scores}
\end{table}

\subsection{Consistency results}
\label{sect:results_consistency}

\begin{table}[t!]
\centering
\begin{tabular}{lcccc}
\toprule
\bf model & \bf deixis & \bf lex. c\!\!. & {\bf ell. infl.}\!\! & {\bf ell. VP}\!\!\!\\ 
 \cmidrule{1-5}
baseline & 50{.}0 & 45{.}9\!\! &  53{.}0\!\! & 28{.}4\!\!\!\\
CADec & 81{.}6 & 58{.}1\!\! & 72{.}2\!\! & \bf{80{.}0}\!\!\!\\
DocRepair & \bf{91{.}8} & \bf{80{.}6}\!\! & \bf{86{.}4}\!\! & 75{.}2\!\!\!\\
\cmidrule{2-5}
 & +10{.}2 & +22{.}5\!\! & +14{.}4\!\! & -4{.}8\!\!\!\\
\bottomrule
\end{tabular}\textbf{}
\caption{Results on contrastive test sets for specific contextual phenomena (deixis, lexical consistency, ellipsis (inflection), and VP ellipsis). }\label{tab:ctx_all_scores}
\end{table}

\begin{figure}[t!]
\center{\includegraphics[scale=0.20]{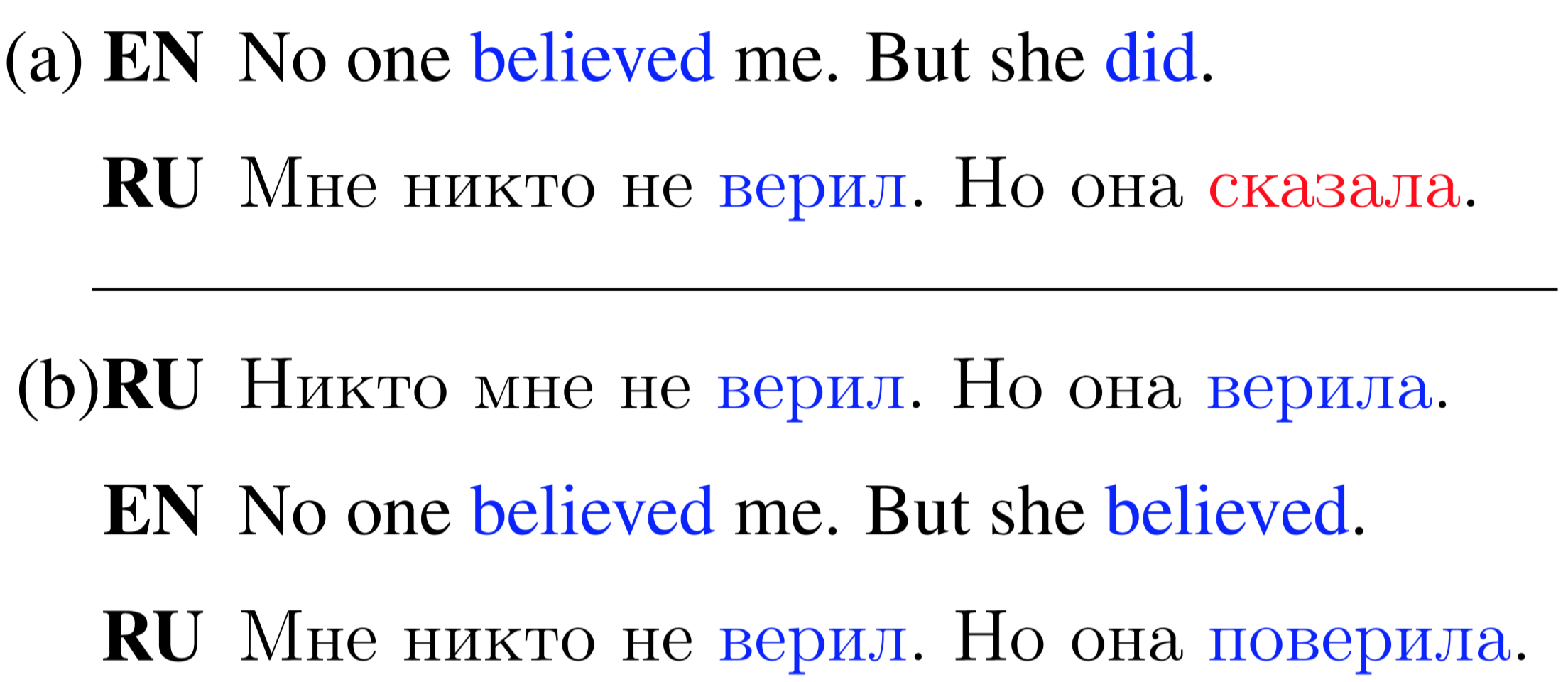}}


\caption{(a) Example of a discrepancy caused by VP ellipsis: correct meaning is ``believe'', but MT produces \begin{otherlanguage}{russian}\emph{сказала} \end{otherlanguage} (``say''). (b) Example of producing round-trip translations. From top to bottom: target, first translation, round-trip translation. When translating from Russian, main verbs are unlikely to be translated into auxiliary ones in English, and VP ellipsis is not present.} \label{fig:ellipsis_vp_example}
\end{figure}

Scores on the phenomena test sets are provided in Table~\ref{tab:ctx_all_scores}. For deixis, lexical cohesion and ellipsis~(infl.) we see substantial improvements over both the baseline and CADec. The largest improvement over CADec (22{.}5 percentage points) is for lexical cohesion. However, there is a drop of almost 5 percentage points for VP ellipsis. We hypothesize that this is because
it is hard to learn to correct inconsistencies in translations caused by VP ellipsis relying on monolingual data alone. Figure~\ref{fig:ellipsis_vp_example}(a) shows an example of inconsistency caused by VP ellipsis in English. There is no VP ellipsis in Russian, and when translating auxiliary ``did'' the model has to guess the main verb. Figure~\ref{fig:ellipsis_vp_example}(b) shows steps of generating round-trip translations for the target side of the previous example. When translating from Russian, main verbs are unlikely to be translated as the auxiliary ``do'' in English, and hence the VP ellipsis is rarely present on the English side. This implies the model trained using the round-trip translations will not be exposed to many VP ellipsis examples in training.
We discuss this further in Section~\ref{sect:varying_data_one_vs_rt}.

Table~\ref{tab:deixis_consistency_scores} provides scores for deixis and lexical cohesion separately for different distances between sentences requiring consistency. It can be seen, that the performance of DocRepair degrades less than that of CADec when the distance between sentences requiring consistency gets larger.

\begin{table}[t!]
\centering
\begin{tabular}{lcccc}
\toprule
&   & \multicolumn{3}{c}{distance}\\
& total & 1 & 2 & 3 \\ 
 \cmidrule{1-5}
\multicolumn{3}{l}{\bf deixis}\\
baseline   &  50{.}0 & 50{.}0 & 50{.}0 & 50{.}0\\
CADec 
& 81{.}6 & 84{.}6 & 84{.}4 & 75{.}9\\
DocRepair & \bf 91{.}8 & \bf 94{.}8 & \bf 93{.}1 & \bf 87{.}7 \\
\cmidrule{2-5}
 & + 10{.}2 & +10{.}2 & +8{.}7 & +11{.}8 \\
\cmidrule{1-5}
\multicolumn{3}{l}{\bf lexical cohesion}\\
baseline   &  45{.}9 & 46{.}1 & 45{.}9 & 45{.}4\\
CADec 
&  58{.}1 &  63{.}2 &  52{.}0 &  56{.}7\\
DocRepair & \bf 80{.}6 & \bf 83{.}0 & \bf 78{.}5 & \bf 79{.}4 \\
\cmidrule{2-5}
& + 22{.}5 & +20{.}2 & +26{.}5 & +22{.}3 \\
\bottomrule
\end{tabular}\textbf{}
\caption{Detailed accuracy on deixis and lexical cohesion test sets. 
} \label{tab:deixis_consistency_scores}
\end{table}

\subsection{Human evaluation}
\label{sect:human_evaluation}

\begin{table}[t!]
\centering
\begin{center}
\begin{tabular}{|c|c|cc|}
\hline
\bf all & \bf equal & \bf better & \bf worse \\
\hline
700 & 367 & 242 & 90 \\ \hline
100\% & 52\% & 35\% & 13\% \\
\hline
\end{tabular}
\end{center}
\vspace{-1ex}
\caption{\label{tab:human_evaluation} Human evaluation results, comparing DocRepair with baseline.}
\end{table}

We conduct a human evaluation on  random 700 examples from our general test set. We picked only examples where a DocRepair translation is not a full copy of the baseline one.\footnote{As we further discuss in Section~\ref{sect:learning_dynamics}, DocRepair does not change the base translation at all in about $20\%$ of the cases.}

The annotators were provided an original group of sentences in English and two translations: baseline context-agnostic one and the one corrected by the DocRepair model. Translations were presented in random order with no indication which model they came from. The task is to pick one of the three options: (1) the first translation is better, (2) the second translation is better, (3) the translations are of equal quality. The annotators were asked to avoid the third answer if they are able to give preference to one of the translations. No  other guidelines were given.

The results are provided in Table~\ref{tab:human_evaluation}. In about $52\%$ of the cases annotators marked translations as having equal quality. Among the cases where one of the translations was marked better than the other, the DocRepair translation was marked better in $73\%$ of the cases. This shows a strong preference of the annotators  for corrected translations over the baseline ones.

\section{Varying Training Data}
\label{sect:varying_data}

In this section, we discuss the influence of the training data chosen for document-level models. In all experiments, we used the DocRepair model.

\subsection{The amount of training data}
\label{sect:amount_of_training_data}

Table~\ref{tab:docrepair_amount_of_data_all_scores} provides BLEU and consistency scores for the DocRepair model trained on different amount of data. We see that even when using a dataset of moderate size (e.g., 5m fragments) we can achieve performance comparable to the model trained on a large amount of data (30m fragments). 
Moreover, we notice that deixis scores are less sensitive to the amount of training data than lexical cohesion and ellipsis scores. The reason might be that, as we observed in our previous work~\cite{voita-etal-2019-good}, inconsistencies in translations due to the presence of deictic words and phrases are more frequent in this dataset than other types of inconsistencies. Also, as we show in Section~\ref{sect:learning_dynamics}, this is the phenomenon the model learns faster in training.

\begin{table}[t!]
\centering
\begin{tabular}{lcccc}
\toprule
 & \bf{BLEU} & \bf deixis & \bf lex. c. & {\bf ellipsis}\\ 
 \cmidrule{1-5}
2{.}5m & 34{.}15 & 89{.}2 & 75{.}5 & 81{.}8 / 71{.}6\\
5m & 34{.}44 & 90{.}3 & 77{.}7 & 83{.}6 / 74{.}0\\
30m & \bf{34{.}60} & \bf{91{.}8} & \bf{80{.}6} & \bf{86{.}4 / 75{.}2}\\
\bottomrule
\end{tabular}\textbf{}
\caption{Results for DocRepair trained on different amount of data. For ellipsis, we show inflection/VP scores. }\label{tab:docrepair_amount_of_data_all_scores}
\end{table}

\subsection{One-way vs round-trip translations}
\label{sect:varying_data_one_vs_rt}

In this section, we discuss the limitations of using only monolingual data to model inconsistencies between sentence-level translations.
In Section~\ref{sect:results_consistency} we observed a drop in performance on VP ellipsis for DocRepair compared to CADec, which was trained on parallel data. We hypothesized that this is due to the differences between one-way and round-trip translations, and now we test this hypothesis.
 To do so, we fix the dataset and vary the way in which the input for DocRepair is generated: round-trip or one-way translations. The latter assumes that document-level data is parallel, and translations are sampled from the source side of the sentences in a group rather than from their back-translations. For parallel data, we take 1{.}5m parallel instances which were used for CADec training and add 1m instances from our monolingual data. For segments in the parallel part, we either sample translations from the source side or use round-trip translations. The results are provided in Table~\ref{tab:round_trip_vs_doc}.

\begin{table}[t!]
\centering
\begin{tabular}{lcccc}
\toprule
\!\! \bf data & \bf deixis & \bf lex. c.\!\! & {\bf ell. infl.} & {\bf ell. VP}\!\!\!\\ 
 \cmidrule{1-5}
\!\! one-way & 85{.}4 & 63{.}4\!\! &  79{.}8 & 73{.}4\!\!\!\\
\!\! round-trip & 84{.}0 & 61{.}7\!\! & 78{.}4 & 67{.}8\!\!\!\\
\bottomrule
\end{tabular}\textbf{}
\caption{Consistency scores for the DocRepair model trained on 2{.}5m instances, among which 1{.}5m are parallel instances. Compare round-trip and one-way translations of the parallel part.}\label{tab:round_trip_vs_doc}
\end{table}

The model trained on one-way translations is slightly better than the one trained on round-trip translations. As expected, VP ellipsis is the hardest phenomena to be captured using round-trip translations,
and the DocRepair model trained on one-way translated data gains 6\% accuracy on this test set. This shows that the DocRepair model benefits from having access to non-synthetic English data. This results in exposing DocRepair at training time to Russian translations which suffer from the same inconsistencies as the ones it will have to correct at test time.

\begin{table}[t!]
\centering
\begin{tabular}{lcccc}
\toprule
\bf data & \bf BLEU &\!\!{\bf deixis}\!\!& \bf lex. c. & {\bf ellipsis}\\ 
 \cmidrule{1-5}
\!\!from mon.\!\!\!\!& 34{.}15\!\!& 89{.}2 & 75{.}5 & 81{.}8 / 71{.}6\\
\!\!from par.\!\!& 33{.}70\!\!& 84{.}0 & 61{.}7 & 78{.}4 / 67{.}8\\
\bottomrule
\end{tabular}\textbf{}
\caption{DocRepair trained on 2{.}5m instances, either randomly chosen from monolingual data or from the part where each utterance in a group has a translation.}
\label{tab:data_mono_vs_doc}
\end{table}

\subsection{Filtering: monolingual (no filtering) or parallel}
\label{sect:amount_of_training_data}

Note that the scores of the DocRepair model trained on 2{.}5m instances randomly chosen from monolingual data (Table~\ref{tab:docrepair_amount_of_data_all_scores}) are different from the ones for the model trained on 2{.}5m instances
combined from parallel and monolingual data (Table~\ref{tab:round_trip_vs_doc}). 
 For convenience, we show these two in Table~\ref{tab:data_mono_vs_doc}. 

The domain, the dataset these two data samples were gathered from, and the way we generated training data for DocRepair (round-trip translations) are all the same. The only difference lies in how the data was filtered. For parallel data, 
as in the previous work~\cite{voita18},
we picked only sentence pairs with large relative time overlap of subtitle frames between source-language and target-language subtitles. This is necessary to ensure the quality of translation data: one needs groups of consecutive sentences in the target language where every sentence has a reliable translation.

Table~\ref{tab:data_mono_vs_doc} shows that the quality of the model trained on data which came from the parallel part is worse than the one trained on monolingual data. This indicates that requiring each sentence in a group to have a reliable translation changes the distribution of the data, which might be not beneficial for translation quality and provides extra motivation for using monolingual data.

\section{Learning Dynamics} 
\label{sect:learning_dynamics}

\begin{figure}[t!]
    \centering
    \begin{subfigure}[b]{0.25\textwidth}
        \includegraphics[width=\textwidth]{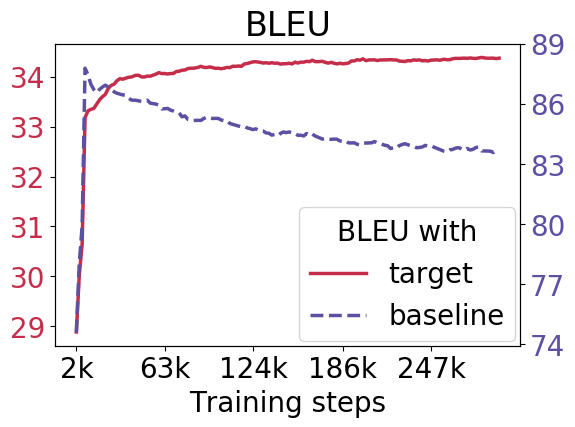}
        \caption{}
        \label{fig:bleu_in_training}
    \end{subfigure}
    \quad
    \begin{subfigure}[b]{0.18\textwidth}
        \includegraphics[width=\textwidth]{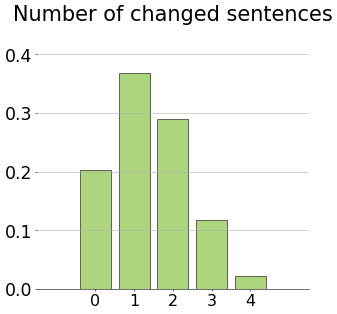}
        \caption{}
        \label{fig:number_of_changed_sentences}
    \end{subfigure}
    \caption{(a) BLEU scores progression in training. BLEU evaluated with the target translations and with the context-agnostic baseline translations (which DocRepair learns to correct). (b) Distribution in the test set of the number of changed sentences in 4-sentence fragments.}
\end{figure}

\begin{figure*}[t!]
\center{\includegraphics[scale=0.50]{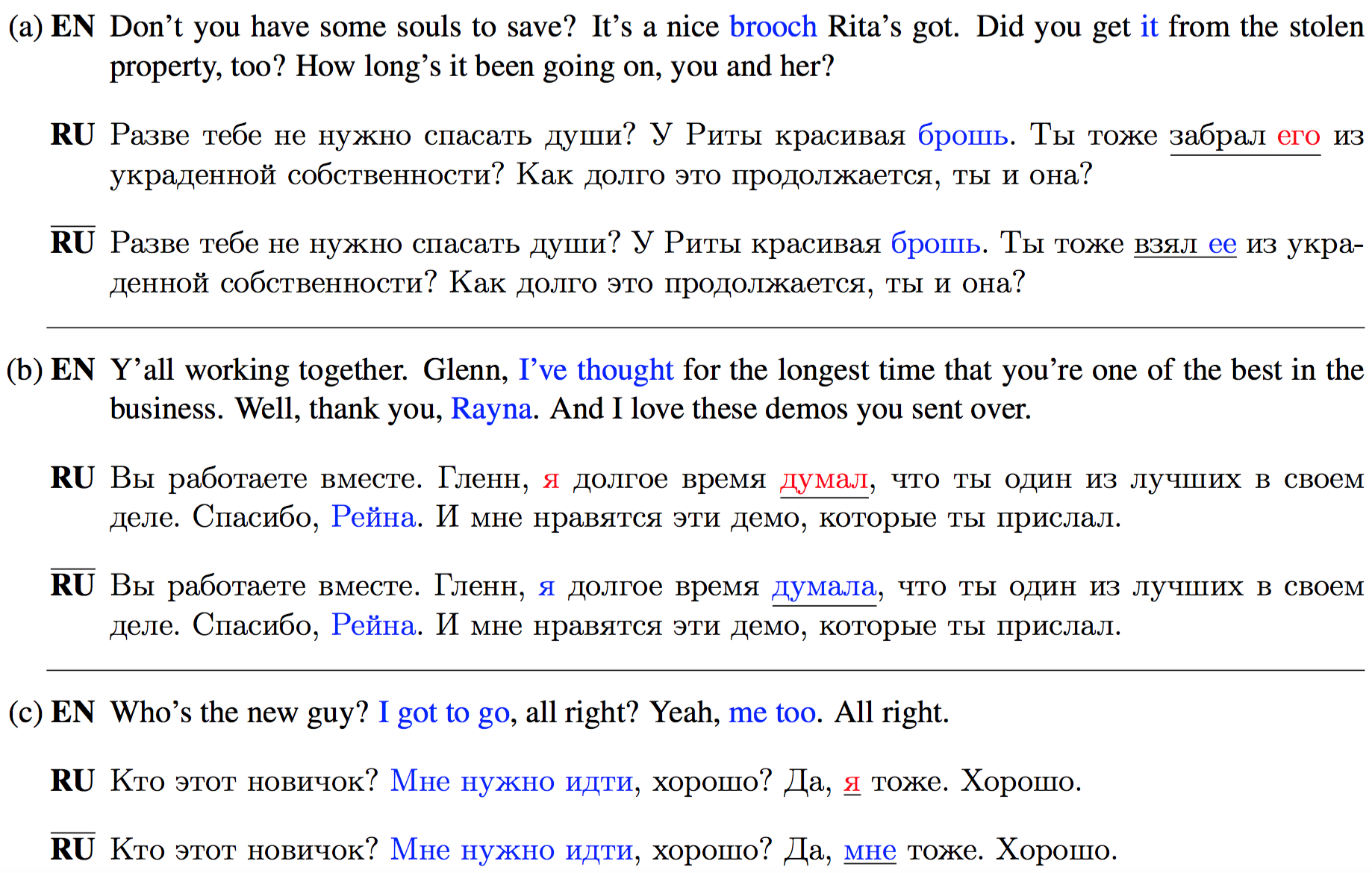}}

\caption{Examples of the DocRepair translations. First is the baseline translation, then -- corrected by the DocRepair. The differences between translations are underlined. (a) corrected wrong translation of ``it'': violated gender agreement with the antecedent. (b) corrected wrong gender (marked on a verb): from the third sentence it's clear that the speaker in the second one is feminine (Rayna), but the baseline translation was masculine. (c) corrected wrong morphological form of the pronoun, which was not understood with the elided verb in the third sentence.}
 \label{fig:translation_examples}
\end{figure*}

Let us now look into how the process of DocRepair training progresses. Figure~\ref{fig:bleu_in_training} shows how the BLEU scores with the reference translation and with the baseline context-agnostic translation (i.e. the input for the DocRepair model) are changing during training. 
First, the model quickly learns to copy baseline translations: the BLEU score with the baseline is very high. Then it gradually learns to change them, which leads to an improvement in BLEU with the reference translation and a drop in BLEU with the baseline. Importantly, the model is reluctant to make changes: the BLEU score between translations of the converged model and the baseline is 82{.}5.
We count the number of changed sentences in every 4-sentence fragment in the test set and plot the histogram in
Figure~\ref{fig:number_of_changed_sentences}.
 In over than 20$\%$ of the cases the model has not changed base translations at all. In almost $40\%$, it modified only one sentence and left the remaining 3 sentences unchanged. The model changed more than half sentences in a group in only $14\%$ of the cases. Several examples of the DocRepair translations are shown in Figure~\ref{fig:translation_examples}.

\begin{figure}[t!]
\center{\includegraphics[scale=0.35]{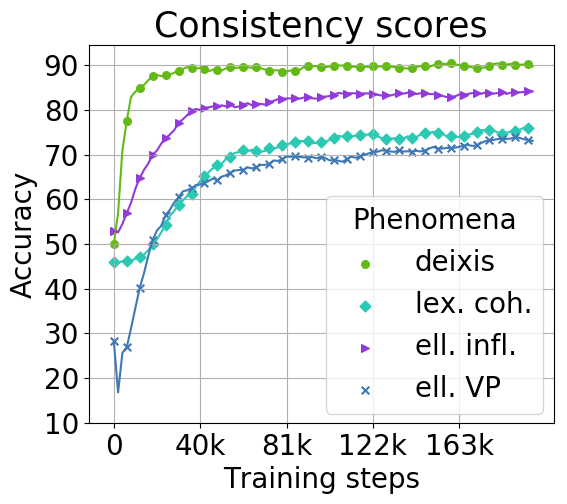}}
\caption{Consistency scores progression in training.}
\label{fig:consistency_in_training}
\end{figure}

Figure~\ref{fig:consistency_in_training} shows how consistency scores are changing in training.\footnote{Deixis and lexical cohesion scores are evaluated on the development sets which were used in training for the stopping criteria. Ellipsis test sets were not used at training time; the scores are shown here only for visualization purposes.} For deixis, the model achieves the final quality quite quickly; for the rest, it needs a large number of training steps to converge.

\section{Related Work}
\label{sect:related_work}

Our work is most closely related to two lines of research: automatic post-editing (APE) and document-level machine translation.

\subsection{Automatic post-editing}

Our model can be regarded as an automatic post-editing system -- a system designed to fix systematic MT errors that is decoupled from the main MT system.
Automatic post-editing has a long history, including rule-based~\cite{Knight:1994:APD:199288.199457}, statistical~\cite{simard-etal-2007-statistical} and neural approaches~\cite{junczys-dowmunt-grundkiewicz-2016-log,pal-etal-2016-neural,freitag2019text}.

In terms of architectures, modern approaches use neural sequence-to-sequence models, either multi-source architectures that consider both the original source and the baseline translation \cite{junczys-dowmunt-grundkiewicz-2016-log,pal-etal-2016-neural}, or monolingual repair systems, as in \citet{freitag2019text}, which is concurrent work to ours. 
True post-editing datasets are typically small and expensive to create \cite{specia2017},
hence synthetic training data has been created that uses original monolingual data as output for the sequence-to-sequence model, paired with an automatic back-translation \cite{sennrich-etal-2016-improving} and/or round-trip translation as its input(s) \cite{junczys-dowmunt-grundkiewicz-2016-log,freitag2019text}.

While previous work on automatic post-editing operated on the sentence level, the main novelty of this work is that our DocRepair model operates on groups of sentences and is thus able to fix consistency errors caused by the context-agnostic baseline MT system.
We consider this strategy of sentence-level baseline translation and context-aware monolingual repair attractive when 
 parallel document-level data is scarce.

For training, the DocRepair model only requires monolingual document-level data.
While we create synthetic training data via round-trip translation similarly to earlier work \cite{junczys-dowmunt-grundkiewicz-2016-log,freitag2019text}, note that we purposefully use sentence-level MT systems for this to create the types of consistency errors that we aim to fix with the context-aware DocRepair model.
Not all types of consistency errors that we want to fix emerge from a round-trip translation, so access to parallel document-level data can be useful (Section~\ref{sect:varying_data_one_vs_rt}).

\subsection{Document-level NMT} 

Neural models of MT that go beyond the sentence-level are an active research area~\cite{jean_does_2017,wang_exploiting_2017,tiedemann_neural_2017,bawden2017,voita18,maruf-haffari:2018:Long,agrawal2018,miculicich-EtAl:2018:EMNLP,kuang-etal-2018-modeling,voita-etal-2019-good}.
Typically, the main MT system is modified to take additional context as its input.
One limitation of these approaches is that they assume that parallel document-level training data is available.

Closest to our work are two-pass models for document-level NMT~\cite{xiong-coherence-delib-2018,voita-etal-2019-good}, where a second, context-aware model takes the translation and hidden representations of the sentence-level first-pass model as its input.
The second-pass model can in principle be trained on a subset of the parallel training data \cite{voita-etal-2019-good}, somewhat relaxing the assumption that all training data is at the document level.

Our work is different from this previous work in two main respects. Firstly, we show that consistency can be improved with only monolingual document-level training data. Secondly, the DocRepair model is decoupled from the first-pass MT system, which improves its portability.

\section{Conclusions}
\label{sect:conclusions}

We introduce the first approach to context-aware machine translation using only monolingual document-level data. We propose a monolingual DocRepair model to correct inconsistencies between sentence-level translations. The model performs automatic post-editing on a sequence of sentence-level translations, refining translations of sentences in context of each other. Our approach results in substantial improvements in translation quality as measured by BLEU, targeted contrastive evaluation of several discourse phenomena and human evaluation.  Moreover, we perform error analysis and detect which discourse phenomena are hard to capture using only monolingual document-level data. While in the current work we used text fragments of 4 sentences, in future work we would like to consider longer contexts.

\section*{Acknowledgments}
We would like to thank the anonymous reviewers for their comments.
The authors also thank David Talbot and Yandex Machine Translation team for helpful discussions and inspiration. Ivan Titov acknowledges support of the European Research Council (ERC StG BroadSem 678254) and the Dutch National Science Foundation (NWO VIDI 639.022.518). 
Rico Sennrich acknowledges support from the Swiss National Science Foundation (105212\_169888), the European Union’s Horizon 2020 research and innovation programme (grant agreement no 825460), and the Royal Society (NAF\textbackslash R1\textbackslash 180122).

\bibliography{emnlp-ijcnlp-2019}
\bibliographystyle{acl_natbib}

\end{document}